\documentclass[letterpaper, 10 pt, journal, twoside]{ieeetran}

\IEEEoverridecommandlockouts                              

\usepackage{amsmath} 
\usepackage{amssymb}  
\usepackage{graphicx}

\usepackage{hyperref}
\usepackage{cleveref} 
\usepackage{rotating} 
\usepackage{booktabs} 
\usepackage{multirow} 
\usepackage{graphicx}
\usepackage{xcolor}
\usepackage{cite}
\usepackage{svg}
\usepackage{color,soul}
\usepackage{enumitem}
\usepackage[switch]{lineno}
\usepackage{fancyhdr}

\title{Visual Sculpting: Visually-Aligned Planning Representations for Long-Horizon Robot Clay Sculpting
}

\author{Peter Schaldenbrand$^{1}$ and Jean Oh$^{1,2}$%
\thanks{\scriptsize $^{1}$The Robotics Institute, Carnegie Mellon University
{\tt\footnotesize pschalde@andrew.cmu.edu jeanoh@cmu.edu}
}%
\thanks{\scriptsize  $^{2}$Lavoro AI Research}
\thanks{\scriptsize \textbf{Accepted for publication in: IEEE Robotics and Automation Letters (RA-L).} Manuscript received: August 25, 2025; Revised: December 3, 2025; Accepted: February 18, 2025.
This paper was recommended for publication by Editor Olivier Stasse upon evaluation of the Associate Editor and Reviewers’ comments.} 
\thanks{\scriptsize  Thank you to Alison Bartsch and Uksang Yoo for helpful advice and to Hyun Woo Park for end-effector designs.
This work was supported in part by NSF IIS-2112633 and JPMorgan Chase Faculty Research Award.}

\thanks{\scriptsize  Digital Object Identifier (DOI): \url{https://doi.org/10.1109/LRA.2026.3673896}}
\thanks{\scriptsize © 2026 IEEE. Personal use of this material is permitted. Permission from IEEE must be obtained for all other uses, in any current or future media, including reprinting/republishing this material for advertising or promotional purposes, creating new collective works, for resale or redistribution to servers or lists, or reuse of any copyrighted component of this work in other works.}
\vspace{-2em}
}

\IEEEpubid{\begin{minipage}{\textwidth}\centering\footnotesize © 2026 IEEE. Personal use of this material is permitted. Permission from IEEE must be obtained for all other uses, in any current or future media, including reprinting/republishing this material for advertising or promotional purposes, creating new collective works, for resale or redistribution to servers or lists, or reuse of any copyrighted component of this work in other works.\\ \textbf{Accepted for publication in: IEEE Robotics and Automation Letters (RA-L). DOI: \url{https://doi.org/10.1109/LRA.2026.3673896}}\end{minipage}}

\begin{document}

\maketitle

\thispagestyle{empty}
\begin{abstract}
    Clay sculpting is a nuanced, artistic task involving dexterous manipulation with long-horizon planning to achieve high-level goals. As a robotics problem, we formulate clay sculpting as a shape-to-shape matching challenge. 
    Prior deformable object manipulation work either requires retraining a policy per goal or relies on dynamics models which represent state as sparse point clouds which do not capture important clay features, such as textures, well.
    We present a method for modeling the dynamics of deformable materials and planning for robotic sculpting in a representation that is visually-aligned, capturing lighting and texture features. 
    With three different deformable materials and various end-effectors, we demonstrate that our dynamics model is comparable in performance to the state-of-the-art with the added benefit of being compatible with visual planning. Our actions are represented as parametrized pushes into clay with a single end-effector, which proved to be suitable for long-horizon ($>$100 actions) clay relief sculptures.
    Lastly, we show the benefits of planning in a visually-aligned representation, but also provide analysis providing evidence as to why this representation is challenging to plan in compared to 3D representations.
\end{abstract}


\vspace{-1em}

\section{Introduction}

\IEEEPARstart{S}{culpting} has a rich history of expressing artistic meanings through 3D forms that captivate our sights, sense of touch, and emotions. Clay sculpting is a long-horizon, dexterous task where an artist takes a sequence of actions to modify the clay until the visual form is aligned with their underlying intentions. In this article, we formulate the process of clay sculpting as a robotic planning problem; that is, given a user's intended goal, a robot takes a sequence of molding actions to create a sculpture matching the goal.

In robotics, clay sculpting is related to deformable object manipulation where the general goal is achieving a target 3D shape. Existing robotic sculpting approaches stemmed from such a 3D shape matching view tend to ignore important visual properties of sculpting, such as textures and shading due to lighting. For instance, subtle changes in textures can create drastic effect for human visual perception of sculptures, but it is hard to measure such an impact when using a 3D metric such as Chamfer Distance on sparse point clouds. To capture such visual guidance as that caused by lighting, we propose a robotic sculpting approach that plans in both a 3D and visually-aligned representations.

\begin{figure}
    \centering
    \includegraphics[width=\linewidth]{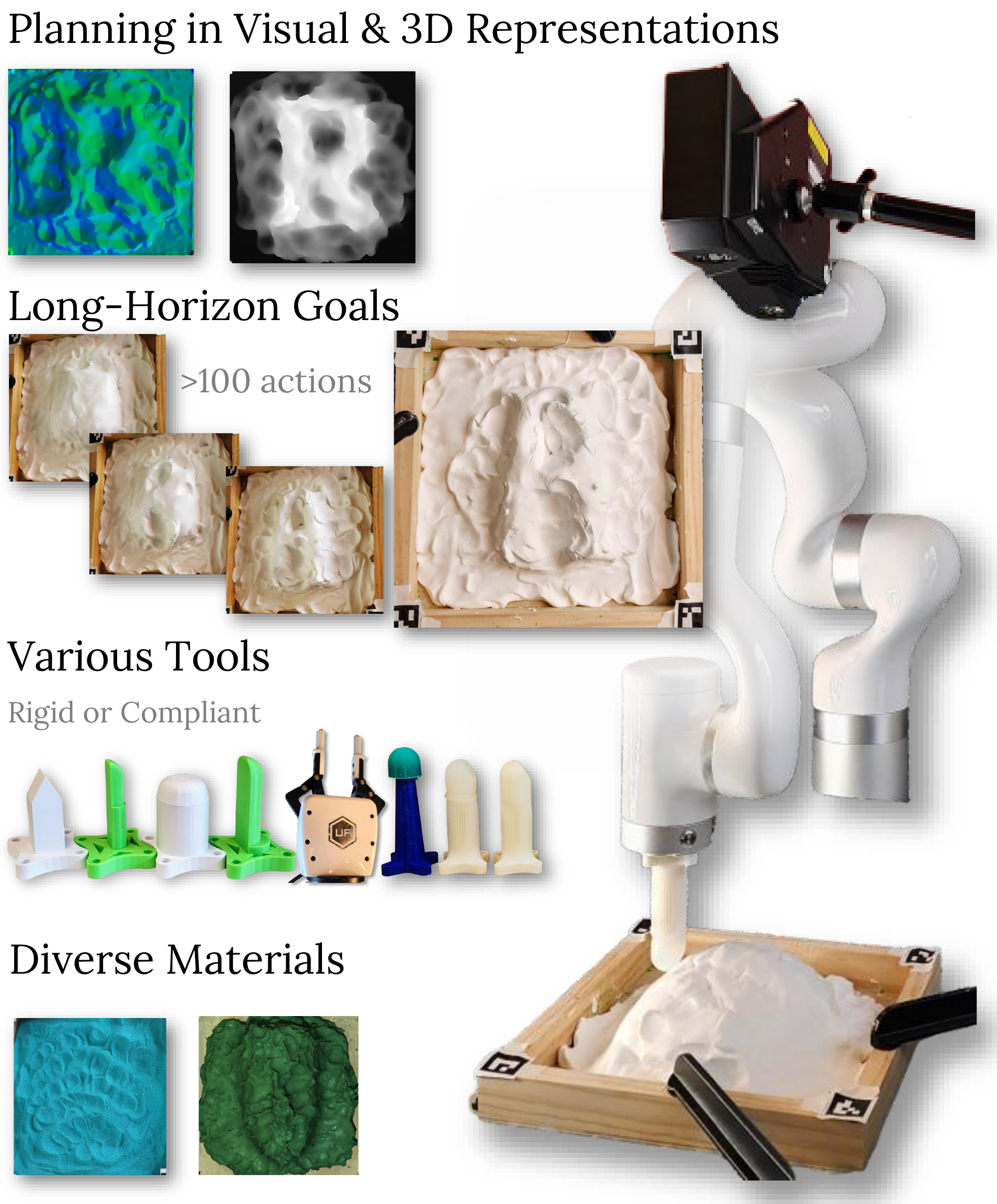}\vspace{-7pt}
    \caption{\textbf{Visual Robotic Sculpting.} We propose an approach to robotic sculpting that models deformable material dynamics in dense, high-resolution depth maps but plans in both 3D and visually-aligned representations in order to more closely align with human perception of 3D objects.}
    \label{fig:figure1}\vspace{-18pt}
\end{figure}

\begin{figure*}
    \centering
    \includegraphics[width=\linewidth]{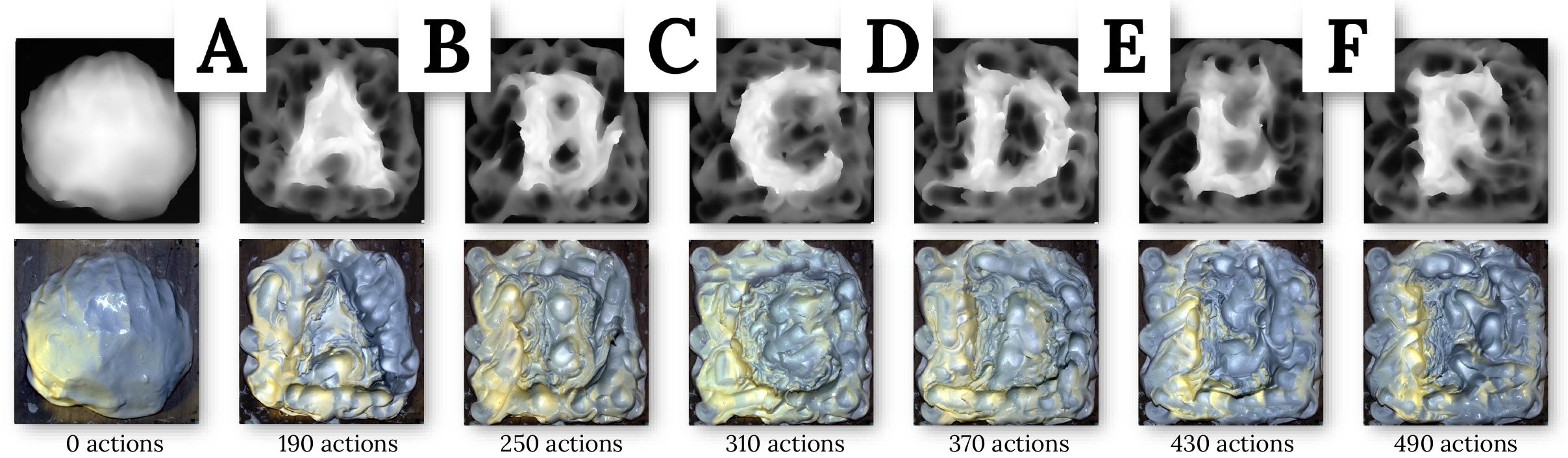}\vspace{-8pt}
    \caption{\textbf{Long-Horizon.} We tested our system's ability to perform long-horizon planning by sculpting the alphabet without resetting the clay between goals. The top row displays the goal images followed by depth maps and photographs of the real sculpted clay along with the total cumulative actions.}
    \label{fig:alphabet} \vspace{-14pt}
\end{figure*}

Although many robotic clay sculpting methods focus on additive or subtractive methods which do not model the dynamics or assume rigidity of the medium~\cite{guljajeva2023psychedelic, duenser2020robocut, brugnaro2018adaptive, ma2020robotsculptor, ma2021stylizedRobotClaySculpt}, there are works that model and embrace the softness and challenges of deformable materials.
Some of these works plan using learning from demonstrations~\cite{bartsch2024sculptdiff, yoo2024ropotter}, but these require retraining policies and recollecting demonstrations for each new goal and starting state. Avoiding this issue, other works model the clay dynamics and plan using policies or Model Predictive Control (MPC)~\cite{shi2023robocook, shi2024robocraft, bartsch2024sculptbot}.

When looking at clay, people do not only see the 3D aspects of the state, they perceive the way light hits the surface and the textures on the clay~\cite{obrien2000texture, todd2004visual}.
Previous work in robotic sculpting model dynamics and plan with sparse ($\sim$300) point clouds~\cite{bartsch2024sculptdiff, shi2024robocraft} which lack a direct visual interpretation and do not capture important features such as texture.
In this paper, we investigate whether a robot can plan to sculpt in both 3D and visually-aligned representations.
We use dense depth maps (512$\times$512) as a 3D representation and the spatial gradient of the depth map as a visually-aligned representation. Spatial gradients capture low-level changes to 3D surfaces and are essential representations used in rendering 3D objects into RGB images, as they are used to estimate the way a given light source interacts and reflects off the surface of the material. Therefore, we consider spatial gradients as a visually-aligned representation, with the assumption that two similar spatial gradients will have similar visual properties.

Clay sculpting is a long-horizon task which may require moving large amounts of material across the working area.
Previous works in deformable object manipulation use pinches as actions using a parallel jaw gripper with 3D printed end-effectors~\cite{shi2023robocook, bartsch2024sculptbot, shi2024robocraft, bartsch2024sculptdiff} which are designed for efficient creation of shapes like alphabet letters. Having two points of contact makes it challenging to make small details, and the pinching motion is not well-suited for moving material across a working area. In this work, we design our actions as simple pushes using a single end-effector to make simple, controlled deformations and better support long-horizon sculpting tasks.

We present a full system for robotic sculpting with visual planning. We represent actions as linear pokes along the surface of the material. We devise a self-supervised data generation scheme and train a dynamics model to predict deformations to depth maps of the material. The dynamics model predictions are differentiably converted to visual representations for planning which is performed using MPC to fit the predicted state to the target state. Comparisons are made both in 3D (e.g., chamfer distance) as well as visual (e.g., mean-squared error of spatial gradients). We show that this leads to sculptures that are not only accurate in 3D but visually similar to target states. 
Our contributions:


\begin{enumerate}
    \item \textbf{A fully-integrated robotic sculpting system} capable of performing long-horizon sculpting due to effective end-effector design (single point of contact versus gripper) and accurate dynamics modeling.
    \item \textbf{A deformable material dynamics model} that is trained from a small set of self-generated actions as a result of our pose-invariant learning mechanism. We show that the model can accurately predict dynamics of diverse materials with varied end-effectors (both soft and rigid).
    \item The first robotic deformable object manipulation planning algorithm which \textbf{plans in both 3D and visually-aligned representations}.
\end{enumerate}

\section{Related Work}

\subsection{Deformable 3D Modeling}

There have been exciting developments in computer-based 3D model generation and manipulation stemming from the rise of large models trained on vast datasets. Many works use 3D representations of Neural Rendering Fields~\cite{poole2022dreamfusion}, Gaussian Splatting, Meshes, or Point Clouds. These works introduce powerful ways to generate 3D models from images, text, or without condition.
These methods provide very powerful tools, such as Score-Distillation Sampling to optimize the shape of a 3D model to fit a given text prompt~\cite{poole2022dreamfusion}, but they are not connected to real-world materials nor have relation to the actions and capabilities of a robot. There are existing material simulators such as PlasticineLab~\cite{huang2021plasticinelab} which use methods such as the Material Point Method (MPM) to estimate the properties of real-world materials, such as clay. Despite improvements in these simulations, previous work has found that simulation methods such as MPM may not perform as well as data-driven approaches (e.g., graph neural networks)~\cite{shi2023robocook, shi2024robocraft}.

\subsection{Robotic Sculpting}

Many robotic sculpting works utilize variants of subtractive or additive actions. Robots have used hot wires to cut through foam~\cite{duenser2020robocut}, loop tools to remove slices of clay~\cite{ma2021stylizedRobotClaySculpt, ma2020robotsculptor, schwartz2013robosculpt}, and chisels to carve wood~\cite{brugnaro2018adaptive}. While these subtractive methods are highly successful at recreating target 3D goals, this success in part comes from the assumption that the materials behave non-deformably. Even subtractive works using clay assume that the tool's path through the clay cleanly removes pieces of clay without creating deformations~\cite{ma2021stylizedRobotClaySculpt, ma2020robotsculptor, schwartz2013robosculpt}. This assumption may work well for hard clay or styrofoam with sharp or hot tools, but will not hold up for very soft materials such as dough or sand.

Prior work has also created sculptures using robotics in an additive manner. These approaches are similar to 3D printing, in which materials are extruded and layered~\cite{guljajeva2023psychedelic, bayram2024clay}. These works do assume deformable properties of the materials, but they heavily engineer the systems to account for this (e.g., extruding very small amounts of clay at a time). These approaches are inherently additive, meaning that they cannot plan to change the existing state of the materials.

\subsection{Robotic Deformable Manipulation}

Rather than assuming rigidity, there are robotic works that can plan with the deformability of materials.
Some works embrace deformability but do not explicitly model dynamics using learning from demonstration~\cite{bartsch2024sculptdiff, yoo2024ropotter} or large language models~\cite{bartsch2024llmCraft, bartsch2025planning} for planning. Other works plan with dynamics models which have been implemented using existing simulators, such as the material point method~\cite{huang2021plasticinelab, shi2024robocraft}, or by training neural networks for a data-driven dynamics modeling~\cite{shi2024robocraft, bartsch2024sculptbot, shi2023robocook}. Although these dynamics models have decreased the Sim2Real gap, they represent clay state as sparse point clouds which do not capture low-level, visual details of clay such as textures. Prior clay manipulation works also operate on small pieces of clay or dough, while in this work our problem setting is to sculpt large surfaces of clay.

Sculpting is long-horizon task; however, most deformable object manipulation works are focused on sculpting simple shapes (e.g., alphabetical letters) from a top-down view, which can be achieved in less than 10 actions. Prior works use parallel jaw grippers with custom end-effectors and represent actions as pinches~\cite{bartsch2024sculptbot, shi2024robocraft, shi2023robocook}. While the pinch action is highly compatible with the alphabetical letter creation task, because of the multiple points of contact and large changes each action makes, it is not suited for long-horizon, fine-grained sculpting such as making relief sculptures with hundreds of actions. In RoPotter~\cite{yoo2024ropotter}, a robot wielding a single end-effector was able to make pots, a task that traditionally uses at least two points of contact.  In this work, we also use pushes with a single end-effector as actions to more closely align with sculpting tasks more generally.

\vspace{-7pt}
\section{Approach}

\vspace{-4pt}
\subsection{Hardware Setup}

Displayed in Fig.~\ref{fig:figure1}, we use the UFactory XArm 850 robot with various, custom end-effectors. The sculpting surface is $12\times12$ inches with Aruco tags on the corners. Suspended directly above the sculpting surface is a Zivid One$^+$ structured light RGB$+$Depth sensor.

\vspace{-4pt}
\subsection{Action Representation}


Robot actions are parameterized as linear pushes with a starting coordinate $(x,y)$; a direction $\theta$; a travel length $l$; and a depth component $z$, as depicted in Fig.~\ref{fig:dynamics-model}.
The action depth is with respect to the surface depth, starting at just lightly touching the surface then pushing $z$ millimeters into the surface by the termination of the trajectory. $z$ and $l$ have maximum values that are set as hyper-parameters.

\vspace{-4pt}
\subsection{End-Effectors}
\begin{figure}
    \centering
    \includegraphics[width=0.9\linewidth]{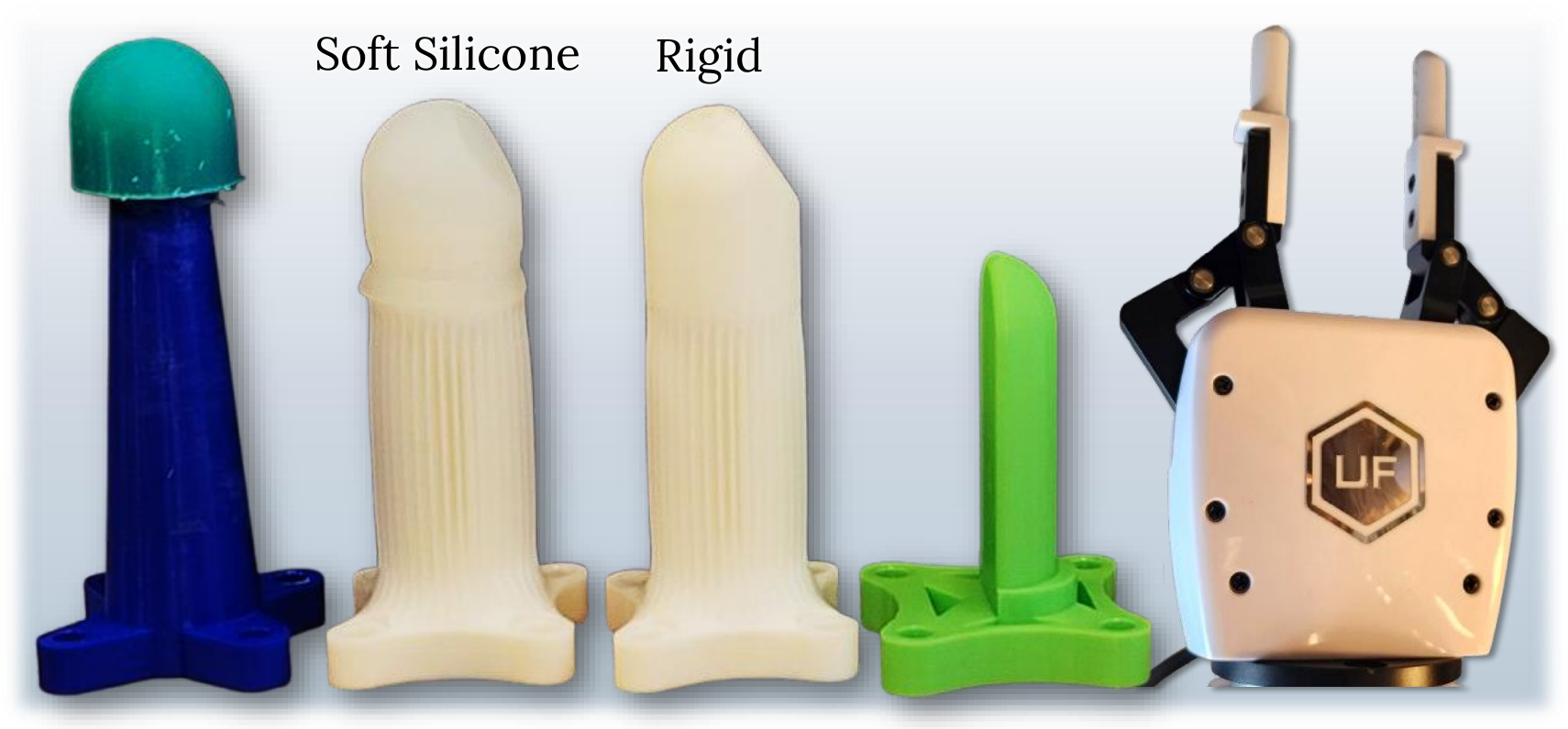}\vspace{-8pt}
    \caption{\textbf{End-Effectors.} - We test our robotic sculpting system with a variety of single end-effectors of various shapes and levels of compliance and compare to a gripper which is conventional in prior work. }\vspace{-14pt}
    \label{fig:end-effectors}
\end{figure}

\begin{figure*}
    \centering
    \includegraphics[width=\linewidth]{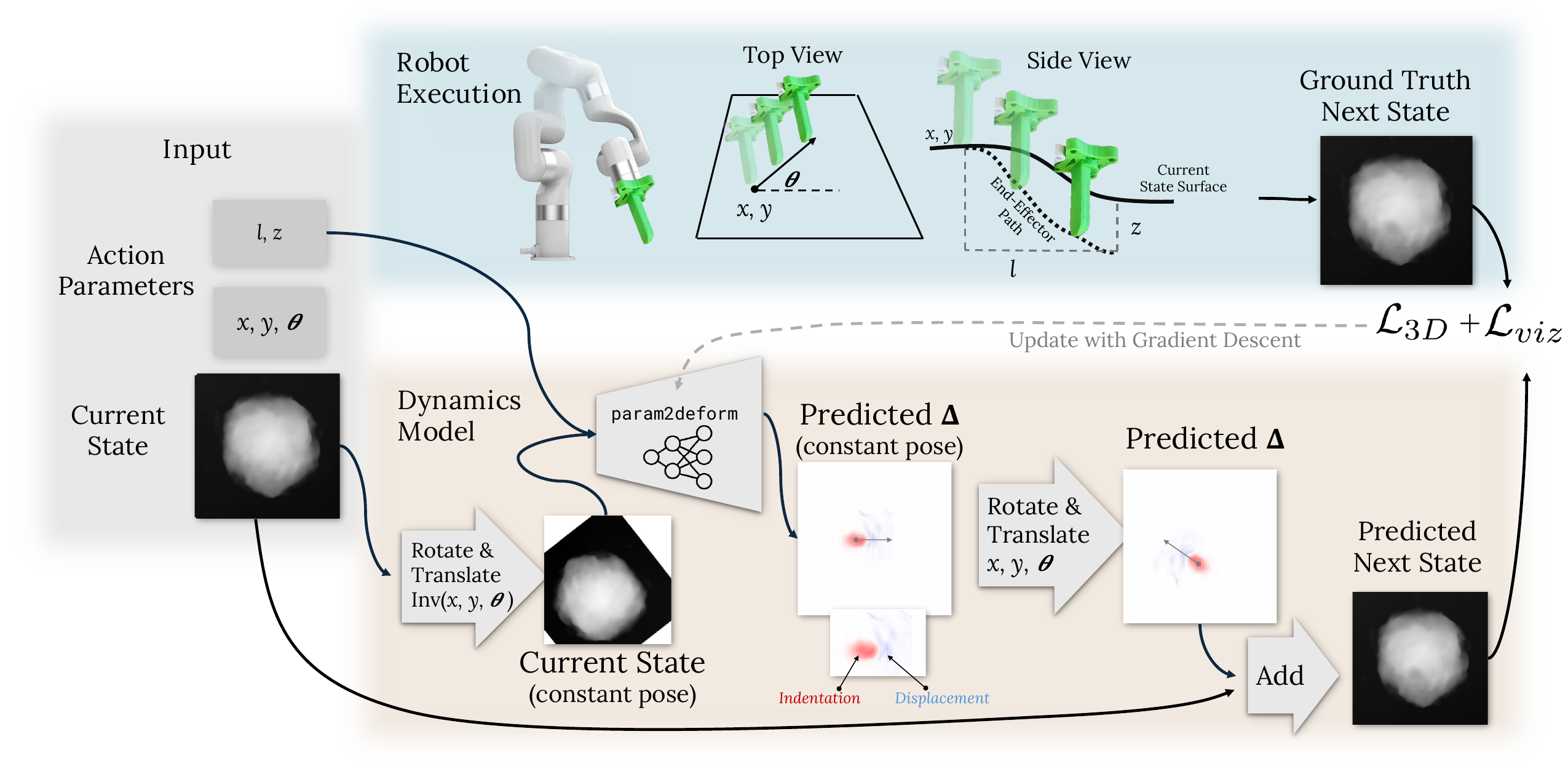}\vspace{-8pt}
    \caption{\textbf{Dynamics Model.} Given the action parameters and current state, our robot can follow trajectories to make deformations along the surface of the material. We model these deformations by training a neural network, \texttt{param2deform}, to predict the changes in state at a constant pose. }
    \label{fig:dynamics-model}\vspace{-14pt}
\end{figure*}

We tested our system with multiple end-effector sizes, shapes, and levels of compliance (Fig.~\ref{fig:end-effectors}).
We chose these end-effectors for their diversity. Our dynamics modeling approach is data-driven, and therefore should adapt to many different shapes and materials of end-effectors without needing prior information, such as 3D model. We also tested a gripper with two custom 3D-printed end-effectors as a baseline comparable to prior works. When using the gripper, we replace the $l$ action parameter with the distance by which the gripper should be closed.

\subsection{Dynamics Model}


The goal of the dynamics model is to predict the change in the material's state given the current state and the action parameters. Our dynamics model is similar to the robot painting system FRIDA~\cite{schaldenbrand2023frida} with notable differences: (1) instead of brush strokes our actions are linear pushes along the material surface, (2) rather than predicting RGB brush stroke appearances our model predict changes to depth maps, and (3) FRIDA assumes that brush stroke actions are independent of the current state, which is not true in sculpting, so our dynamics model includes the current state when predicting changes.

Fig.~\ref{fig:dynamics-model} shows a visual depiction of our dynamics model. At the core of the model is a neural network, \texttt{param2deform}, which predicts the change in depth. \texttt{param2deform} is comprised of 3 multilayer networks. The first takes the two shape parameter scalars, $l$ and $z$, feeds them through 3 linear layers, resizes them to a 2D matrix, then feeds them through 3 convolutional layers. The current state is concatenated with its spatial gradient, then encoded using 3 convolutional layers. Finally, the current state features, current state, and encoding from the shape parameters are concatenated then fed through 5 convolutional layers to predict the final change in state. This network architecture was chosen for its simplicity and hyper-parameters (e.g., number of convolutional layers and hidden layer sizes) were tuned by hand.

Training data is created by uniformly sampling action parameters and recording state before and after the action is taken by the robot. This self-generating data creation phase is performed initially to train the dynamics model. As the robot sculpts, additional training data can be recorded to improve the dynamics model, but for clarity of results, in this paper, the models were trained prior to sculpting.

To reduce the training data needed to learn the model, \texttt{param2deform} predicts all deformations in a constant pose, meaning the start of the action is always at the same point and the action moves from left to right. This predicted deformation is then translated into the desired position (incorporating the $x$, $y$, and $\theta$ parameters) using perspective warps which do not require training data to perform and are fully differentiable.

\subsubsection{Dynamics Model Objectives}

We sample random actions and capture scans of the depth of the materials before ($S_{t}$) and after ($S_{t+1}$) forming training data for \texttt{param2deform}.
We desired for our dynamics model, $f$, to be accurate in both 3D and visual representations, so we optimized \texttt{param2deform} with two different objectives to achieve these goals.
Our 3D loss function, $\mathcal{L}_{3D}$ (Eq.~\ref{eq:loss_3d}), is the mean-squared error between the actual depth map after the action and the dynamics model prediction. To capture the visual features, we form a visual loss function, $\mathcal{L}_{viz}$ (Eq.~\ref{eq:loss_viz}), which is the difference in spatial gradients of the  actual depth map after the action and the dynamics model prediction.

\begin{align}
    \mathcal{L}_{3D} &= ||S_{t+1} - f(S_{t}, a) || \label{eq:loss_3d} \\
    \mathcal{L}_{viz} &= || \nabla S_{t+1} - \nabla f(S_{t}, a) || \label{eq:loss_viz}
\end{align}

We can also convert the depth maps into point clouds and compute more standard loss functions like Chamfer Distance (CD) and Earth-Mover's Distance (EMD). Since the depth maps are high-resolution ($512\times512$), we must down-sample our point clouds for computational purposes. We use voxel-grid down-sampling and the computations of CD and EMD from \cite{shi2023robocook}.



\subsection{Planning}

The goal of our planning algorithm is to recreate a given 3D model in both a 3D and visual representation. The number of actions, starting state, and robot end-effector choice are given. We employ a combination of random-sampling with Gradient Descent for optimizing the action parameters to achieve this goal.

\subsubsection{Goal Creation and Processing} \label{sec:goal-creation}

\begin{figure*}
    \centering
    \includegraphics[width=\linewidth]{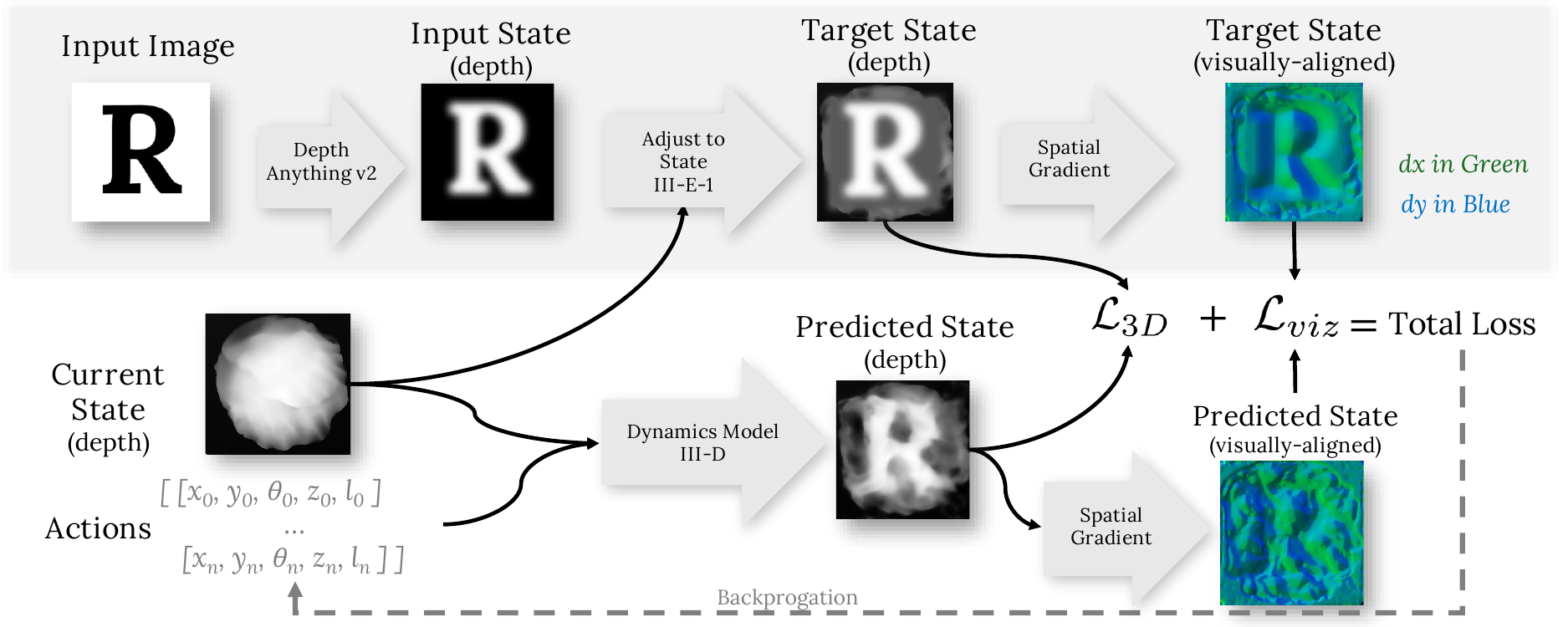}\vspace{-7pt}
    \caption{\textbf{Planning.} (Above) An image is specified by a user and is then converted to depth. The depth map is altered to make it more feasible for the robot to create based on the current state of the material forming a target state. (Below) Our planning algorithm optimizes a set of randomly initialized actions such that the dynamics model predicted state is both accurate in 3D and visual representations compared to the target state.}\vspace{-8pt}
    \label{fig:planning}
\end{figure*}

Our planning algorithm requires a dense, depth map as the target state representation, which can be given directly as a depth map from a 3D model or can be extracted from a given image input by using the pre-trained image-to-depth model, DepthAnythingV2~\cite{depth_anything_v2}.

The input depth map is not calibrated to the given starting state, so, for example, there may not be enough clay in the current state to recreate the target depth map. We adjust the input target depth map ($S_t$) with the current state depth map ($S_0$) with an optimization of the surface. Shown in Eq.~\ref{eq:adjust_input}, we scale the target depth map by $\alpha$ and $\beta$. Because it is difficult for our robot to work near the edges of the working area, we ensure that the edges of the  final target depth, $\hat{S}$, are equal to the current state using boolean map, $M$, which isolates the outer 10\% of the working area.   Our optimization function, Eq.~\ref{eq:adjust_optimization}, optimizes scalars $\alpha$ and $\beta$ such that the target depth map, $\hat{S}$, has (1) the same amount of material as the current state, (2) does not have depth values larger than the table surface, $d_{max}$, and (3) has the most definition (defined as a large $\alpha$ value). These three terms are weighted ($w0$, $w1$, and $w2$) by hand from experimenting with a few test cases.

\begin{align}
    \hat{S} &=   S_0[M] + (\alpha S_t + \beta)[1-M]  \label{eq:adjust_input} \\
    \min_{\alpha, \beta}  &~ w_{0} (\Sigma \hat{S} - \Sigma S_0) + w_{1} \Sigma \hat{S}[\hat{S} > d_{max}] - w_{2} \alpha \label{eq:adjust_optimization}
\end{align}

\subsubsection{Planning Objectives}

Similar to our dynamics model objectives, our planning objective is to recreate both the 3D shape and visual attributes of a given shape model. We  optimize the action parameters, $\mathbf{a}$, to achieve these objectives (Eq.~\ref{eq:planning_objective}). We compare the 3D shape of the target shape, $S^\prime$, and our dynamics model prediction, $f(S_0, \mathbf{a}) $, using mean-squared error matching pixels of depth maps. 

\begin{align}
    \mathcal{L}_{3D} &= ||S^\prime - f(S_0, \mathbf{a}) || \label{eq:loss_3d_planning} \\
    \mathcal{L}_{viz} &= || \nabla S^\prime - \nabla f(S_0, \mathbf{a}) || \label{eq:loss_viz_planning} \\
    \min_{\mathbf{a}} ~ & [ w_{3D} \mathcal{L}_{3D} + w_{viz} \mathcal{L}_{viz} ]  \label{eq:planning_objective}
\end{align}

\subsubsection{Planning Algorithm}

Our planning algorithm is a simple variant of MPC. A given number of actions are initialized using greedy sampling. Actions are initialized one-by-one picking the action that decreases the loss the most over a number of trials. These initialized actions can then be optimized using gradient descent or cross-entropy method to decrease the loss values. While the initialization is greedy, this optimization stage helps promote long-horizon planning since all actions are influential in the objective. This forms an initial plan. A small number of actions are performed, then the robot pauses to update and optimize the remaining plan, to adjust for differences in the dynamics model prediction and reality. This process is repeated until all of the actions in the plan are performed. In practice, we plan with at least 40 actions to help promote long-horizon planning and avoid plans converging on local minima.

\vspace{-4pt}\section{Results}

\subsection{Dynamics Model}

\begin{figure}
    \centering
    \includegraphics[width=\linewidth]{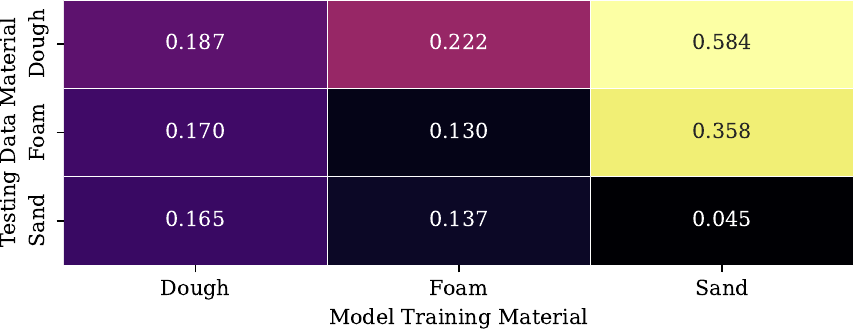}  \vspace{-8pt}
    \caption{\textbf{Out-of-Distribution Dynamics Modeling.} We train our dynamics model on one material and test on another. Reported above are Sim2Real gap values (lower is better) computed as the MSE between predicted and true depth maps (Eq.~\ref{eq:loss_3d}).} 
    \label{fig:dm-cross-materials}  \vspace{-15pt}
\end{figure}

\subsubsection{Qualitative \& Quantitative Results}
\begin{figure}
    \centering
    \includegraphics[width=\linewidth]{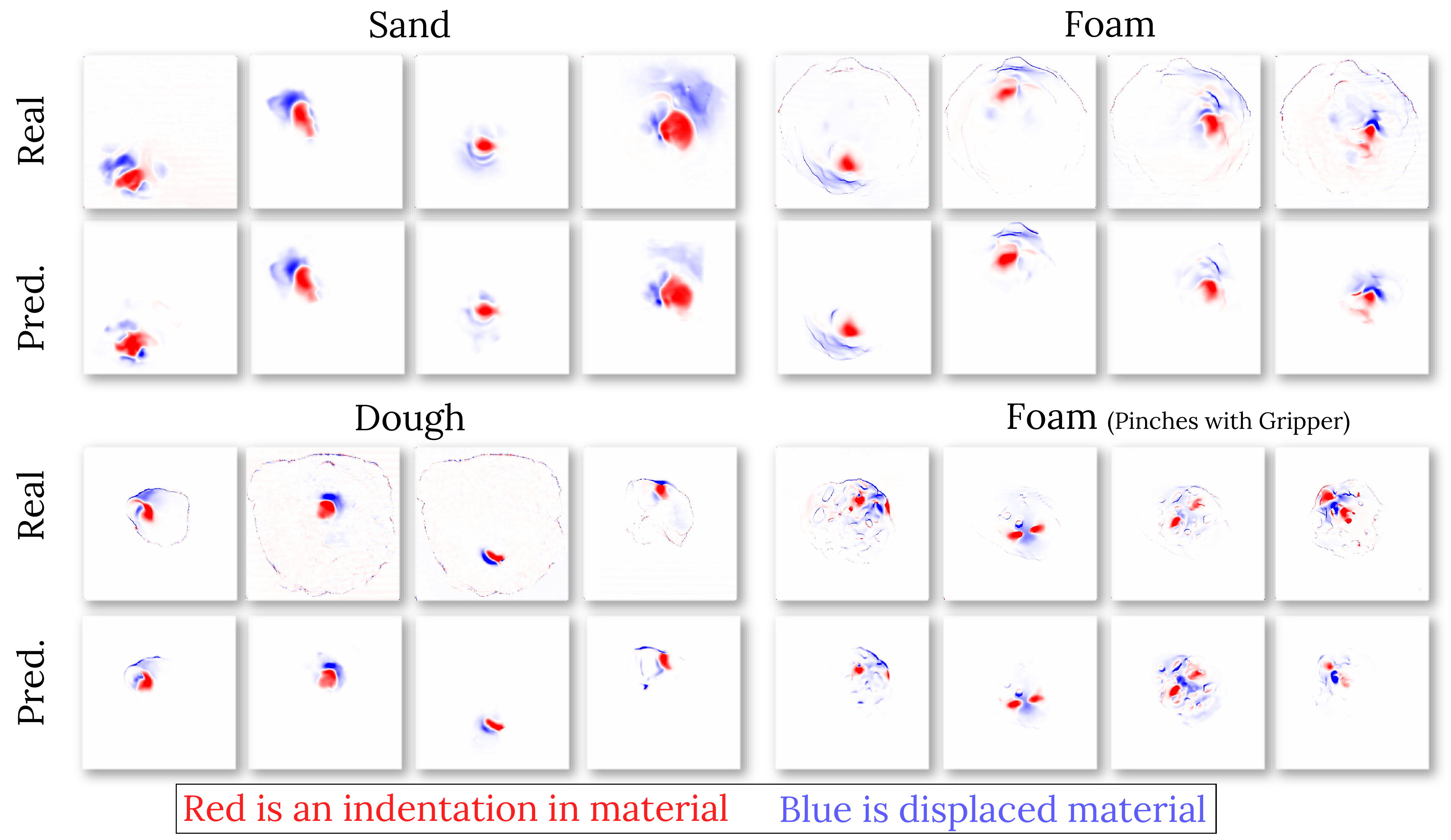}\vspace{-7pt}
    \caption{\textbf{Qualitative Dynamics Model Results.} The top rows show real deformations made into various materials by our robot. Our dynamics model predictions given the current state and action parameters are shown below the real deformations.}
    \label{fig:dm_qualitative}  \vspace{-8pt}
\end{figure}

To test the accuracy of our dynamics model and investigate the effect of our visual loss term, we train our dynamics model with various materials using the same end-effector and show the performance on various metrics on a held-out set of deformations in Table~\ref{tab:dm_viz_3d}. Generally, the addition of the visual loss term, $\mathcal{L}_{viz}$, increased performance not only on the visual evaluation metric but also on the 3D metrics. However, the visual guidance did not improve results for sand. We attribute this to the lack of visual complexity in the sand deformations as compared to the wrinkling and textures created by working with foam and dough as seen in Fig.~\ref{fig:dm_qualitative}.

For qualitative investigation of our dynamics model, we displayed sample predicted deformations in Fig.~\ref{fig:dm_qualitative}. Overall, our model is able to capture the complex deformations in various materials in the local region but fails to predict small deformations far away from the contact. Between the tested materials, foam hosted highly complex deformations while sand had unpredictable deformations that depended greatly on the gradient of the surface (sand rolling down a hill).

\subsubsection{Generalization Across Materials} 
In Fig.~\ref{fig:dm-cross-materials}, we evaluated our dynamics model trained on data from one material and then tested data from another material. We found that some materials lead to good generalization, as a model trained on foam and tested on dough performed only slightly worse than that trained on dough itself. However, training on sand led to very poor generalization to other materials. Overall, these results show that deformable materials are nuanced and cannot be modeled by a single set of parameters.

\begin{table}[]
    \centering 
    \caption{\textbf{Dynamics Modeling.} - Dynamics model performance on a held out set of deformations with various materials while ablating the training objectives. Lower is better for all metrics.}
    \label{tab:dm_viz_3d}
    \begin{tabular}{cc|ccccc} 
        \toprule
        \textbf{} & \textbf{Objective(s)} &  \textbf{$\mathcal{L}_{3D}$} & \textbf{$\mathcal{L}_{viz}$} & \textbf{$CD$} & \textbf{$EMD$} \\
        \midrule
        \multirow{2}{*}{\rotatebox[origin=c]{}{\parbox{2cm}{\centering\textbf{Foam}}}} 
        &  $\mathcal{L}_{3D}$ & 0.138 & 0.025 & 0.26 & 0.16 \\
        & $\mathcal{L}_{3D} +  \mathcal{L}_{viz}$ & \textbf{0.130} & \textbf{0.024} & \textbf{0.22} & \textbf{0.15} \\
        \cmidrule{1-6} 
        \multirow{2}{*}{\rotatebox[origin=c]{}{\parbox{2cm}{\centering\textbf{Sand}}}} 
         & $\mathcal{L}_{3D}$ & \textbf{0.043} & 0.012 & \textbf{0.40} & \textbf{0.22} \\
         & $\mathcal{L}_{3D} +  \mathcal{L}_{viz}$ & 0.047 & \textbf{0.011} & \textbf{0.40} & \textbf{0.22} \\
        \cmidrule{1-6} 
        \multirow{2}{*}{\rotatebox[origin=c]{}{\parbox{2cm}{\centering\textbf{Dough}}}} 
         & $\mathcal{L}_{3D}$ & 0.190 & 0.029 & 0.45 & 0.31 \\
         & $\mathcal{L}_{3D} +  \mathcal{L}_{viz}$ & \textbf{0.187} & \textbf{0.028} & \textbf{0.41} & \textbf{0.30} \\
        \cmidrule{1-6} 
        
        \cmidrule{1-6} 
        
        \multirow{1}{*}{\rotatebox[origin=c]{}{\parbox{2cm}{\centering\textbf{Foam Pinch}}}} 
          & $\mathcal{L}_{3D} +  \mathcal{L}_{viz}$ & 0.624 & 0.043 & 0.50 & 0.30 \\
        \cmidrule{1-6} 
        \bottomrule
    \end{tabular}\vspace{-15pt}
\end{table}

\subsubsection{Dynamics Model Sample Efficiency}
\begin{figure}
    \centering
    \includegraphics[width=\linewidth]{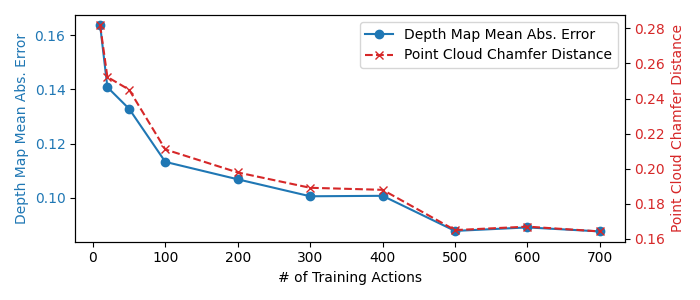}\vspace{-7pt}
    \caption{\textbf{Dynamics Model Sample Efficiency} - Our dynamics model is able to learn an accurate transition model with as few as 100 actions.}
    \label{fig:dm-efficiency}  \vspace{-8pt}
\end{figure}

We trained our dynamics model with varying numbers of training samples and displayed the results in Fig.~\ref{fig:dm-efficiency}. Our model performs well with only roughly 100 samples, but performance increases with more samples.

\subsection{Planning Results}

\subsubsection{Long-Horizon Planning}

To test our method's ability to handle long-horizon tasks, we create a series of goals of alphabetic letters in serif font. The robot first sculpted an ``A" from a starting state, then morphed it into a ``B" and so on.
In Fig.~\ref{fig:alphabet}, we report results to ``F", showing the robot's capability to plan over hundreds of actions.

\subsubsection{Pinching versus Pushing}

We compared our single end-effector pushing actions with a pinching action using a gripper that is analogous to other deformable manipulation works~\cite{shi2023robocook, shi2024robocraft, bartsch2024sculptbot, bartsch2024sculptdiff}. In Table~\ref{tab:dm_viz_3d}, we reported the results of dynamics modeling which were worse for the pinching actions, indicating that pinches produced less predictable deformations than our pushing actions. This was supported by qualitative results in Fig.~\ref{fig:dm_qualitative} which showed that the pinches were complex and not modeled as accurately. In Fig.~\ref{fig:losses-long-horizon}, both visual and 3D losses decreased as many actions were taken using the single end-effector pushes, however, only the 3D losses improved with the gripper pinches. We found that the pinches produced choppy, messy sculptures over the course of many actions.

\subsubsection{Comparison with Other Dough Works}

\begin{figure}
    \centering
    \includegraphics[width=\linewidth]{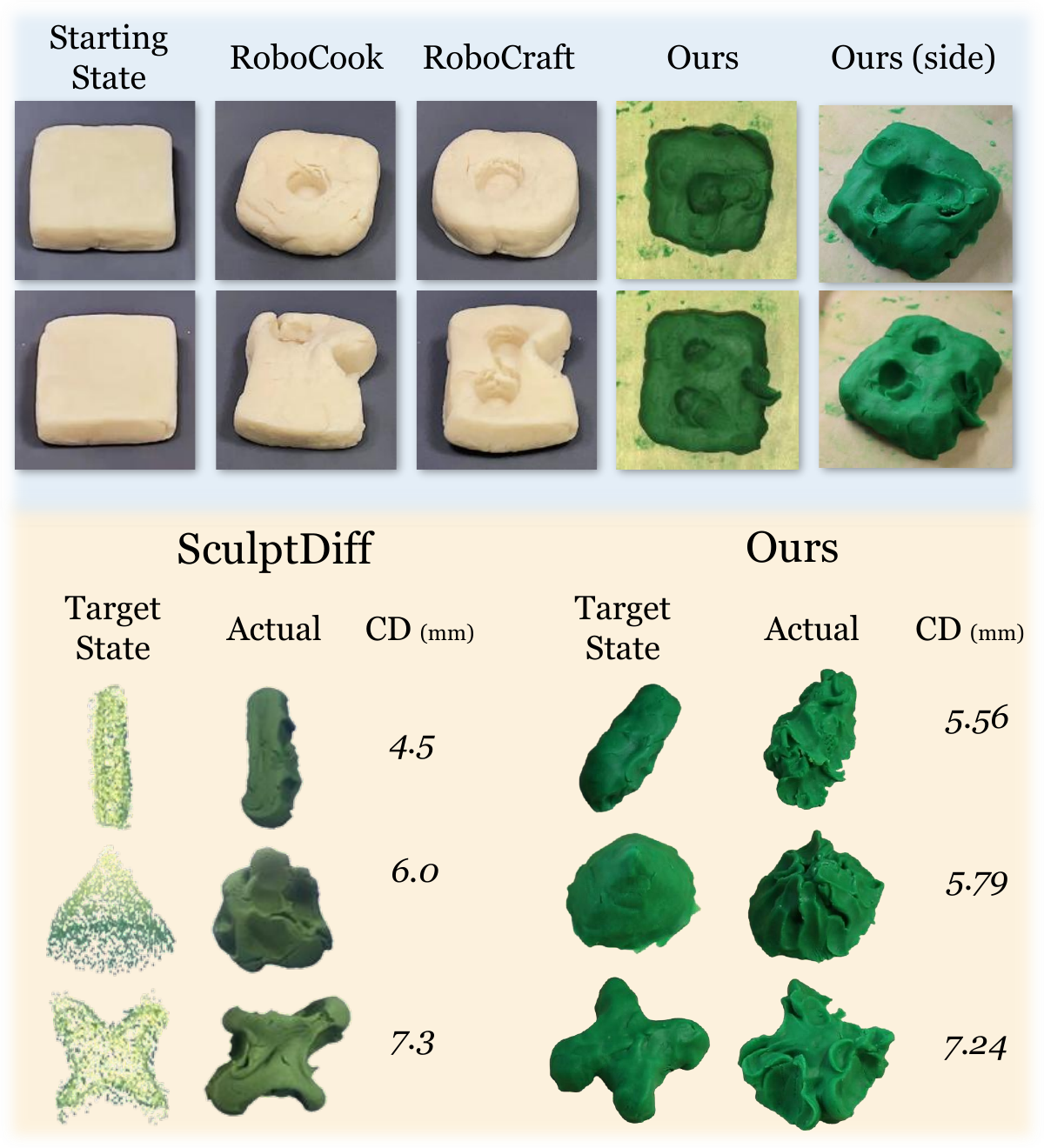}\vspace{-7pt}
    \caption{\textbf{Simple Shape Results} - While our method was designed for large, clay surface sculpting, we compare our results to prior work on simple, small shapes.
    Our method has similar results with simple shapes, such as letters and pyramids, to other play-doh manipulation works, RoboCraft~\cite{shi2024robocraft}, RoboCook~\cite{shi2023robocook}, and SculptDiff~\cite{bartsch2024sculptdiff}. }
    \label{fig:simple-shapes}  \vspace{-14pt}
\end{figure}

In Fig.~\ref{fig:simple-shapes}, we compared our approach to existing deformable object manipulation works. Since we were unable to replicate results from these works, the images of the results were taken from the papers along with the metric results. We attempted to use similar starting and goal states to the compared works. Although our approach was designed for larger, long-horizon sculptures, this comparison served as preliminary evidence that our approach is comparable on the task of making simple, small shapes to existing approaches~\cite{shi2023robocook, shi2024robocraft, bartsch2024sculptdiff}. However, because of the lack of control in this experiment, we are unable to draw broader conclusions about the differences in results.

\subsubsection{Goal Creation and Processing}
\begin{figure}
    \centering
    \includegraphics[width=\linewidth]{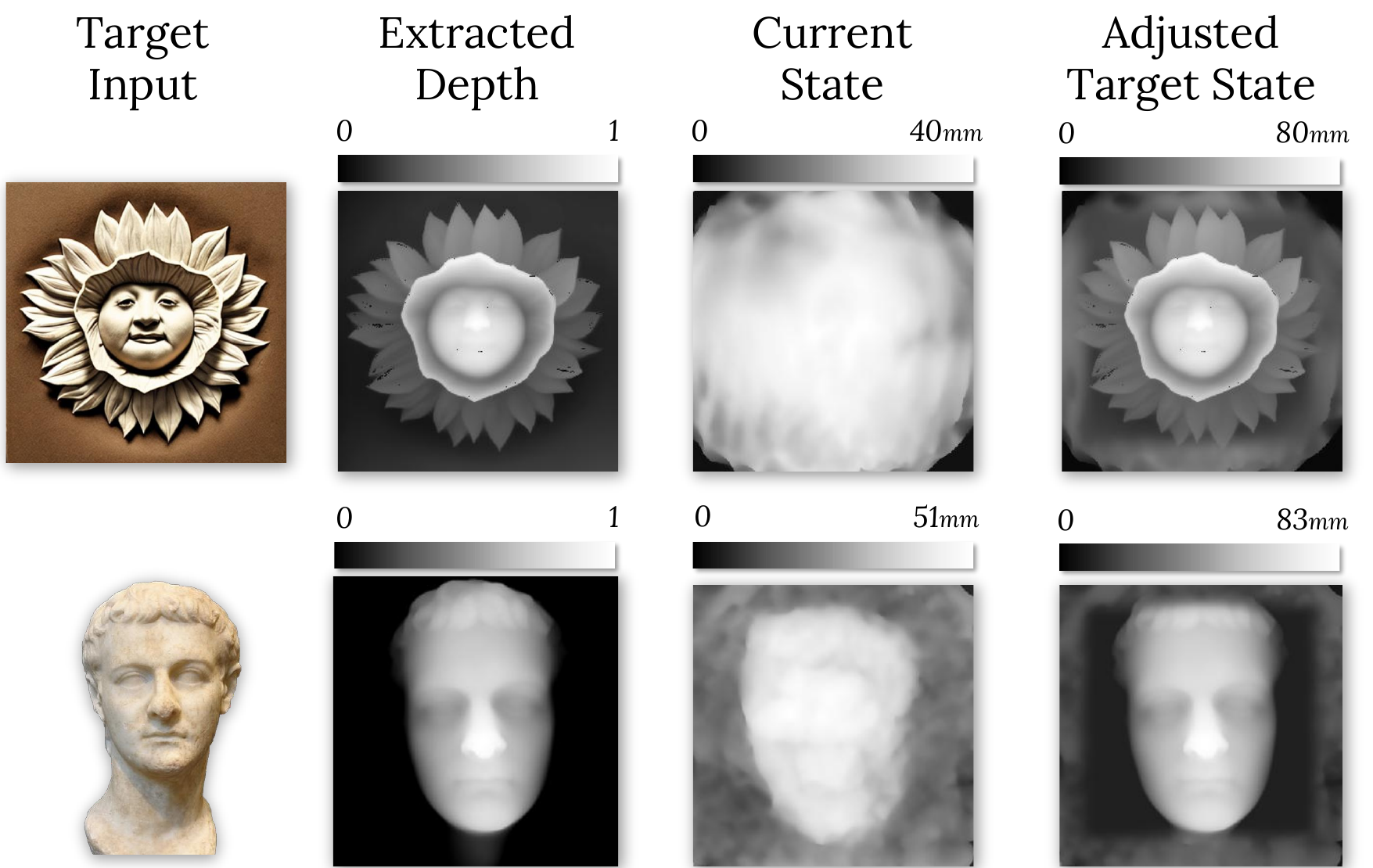}\vspace{-7pt}
    \caption{\textbf{Goal Creation.} Target depth maps are adjusted so that they are more feasible for the robot to recreate. Details in Sec.~\ref{sec:goal-creation}.}
    \label{fig:adjust_goal}\vspace{-7pt}
\end{figure}

In Fig.~\ref{fig:adjust_goal}, we show two examples of the processing steps when receiving an input modality to be used as a target shape. In the upper example, an RGB image is generated by an image generator, then the depth is extracted using DepthAnythingV2~\cite{depth_anything_v2}. Using the current state, this extracted depth is adjusted according to the optimization described in Sec.~\ref{sec:goal-creation}. The second example in Fig.~\ref{fig:adjust_goal}, shows an example where a 3D model downloaded from the Internet is converted to depth, then adjusted according to Sec.~\ref{sec:goal-creation}.

\begin{figure}
    \centering
    \includegraphics[width=\linewidth]{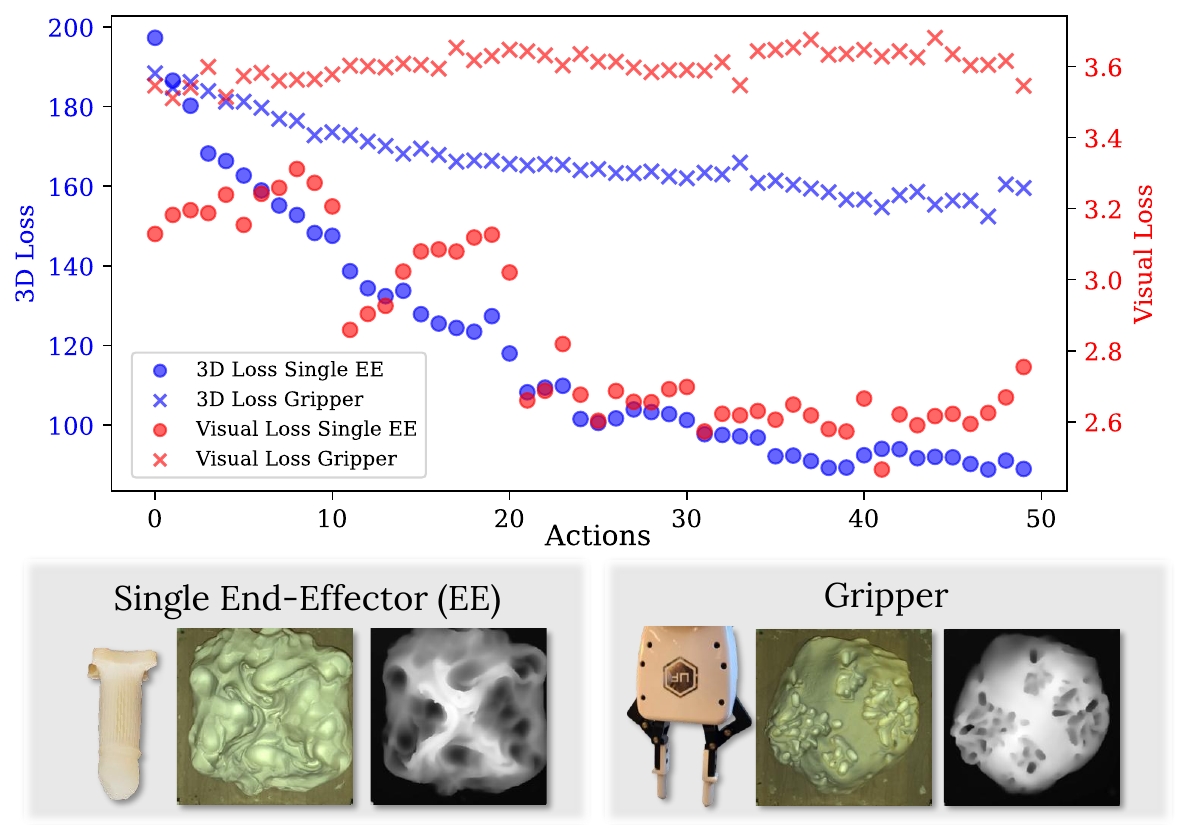} \vspace{-14pt}
    \caption{\textbf{Visual and 3D losses during long-horizon sculpting.} The losses were plotted after each of 50 actions taken by the robot using a single end-effector with our pushing actions and compare to a gripper using pinch actions analagous to prior works~\cite{shi2023robocook, shi2024robocraft, bartsch2024sculptbot, bartsch2024sculptdiff}. Below, we show samples of photographs and depth scans of the material after the actions were taken.}
    \label{fig:losses-long-horizon} \vspace{-15pt}
\end{figure}

\subsubsection{3D-Visio Planning}

To observe our approach's ability to improve 3D and visual accuracy over many actions,
we plotted the losses as actions were taken as the robot sculpted a large ``X" relief sculpture in Fig.~\ref{fig:losses-long-horizon}. Our approach (with a single end-effector) is able to decrease both the visual and 3D losses over many actions, though, it is worth noting that the visual loss did not decrease steadily and leveled off before the 3D loss.

\begin{figure}
    \centering
    \includegraphics[width=\linewidth]{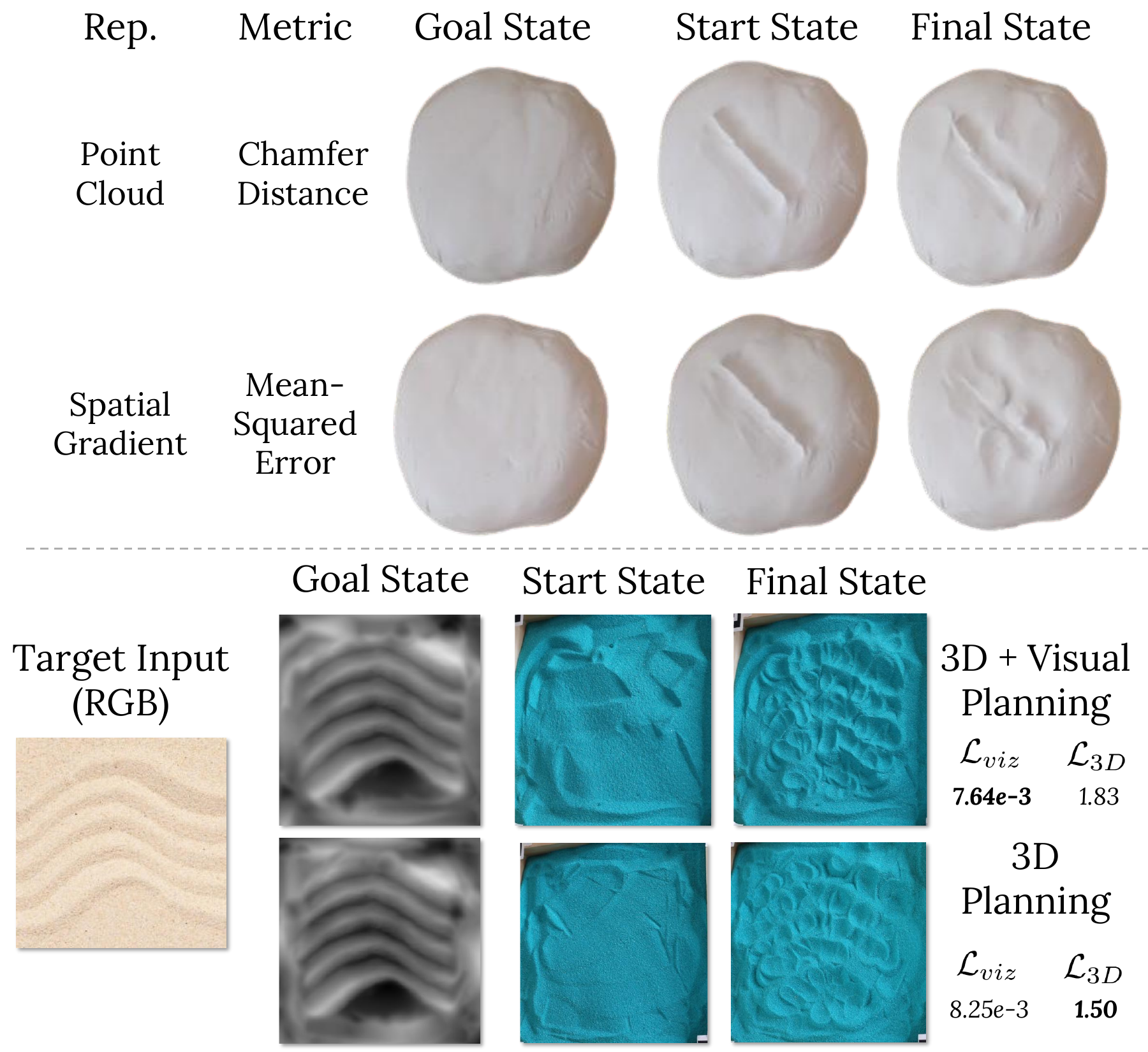}
    \caption{\textbf{Sculpting in a Visual Representation.} (Above) We isolate planning in 3D (minimizing Chamfer Distance) and visual (minimizing mean-squared error of spatial gradients) representations. In both conditions, the robot performed 10 actions to smooth out a line pinched in the clay. 
    Planning in a visual space creates plans that properly align with the task. 
    (Below) When planning with more complex goals with sensitive materials (sand), the effect of visual guidance was not as apparent.}
    \label{fig:cd_vs_sg}  \vspace{-14pt}
\end{figure}

To investigate the effect of 3D and visual losses, we performed an ablation study of the losses in the planning objective and reported the results in Fig.~\ref{fig:cd_vs_sg}. We created a simple case where the robot's objective was to smooth out a thin line pinched into the material. This example showed an extreme change in visual representation, whereas the change in 3D was not as extreme because the line was very thin. When planning with a point cloud representation and chamfer distance, the robot did not smooth out the line well, whereas when planning with visual loss, most of the robot's actions worked to smooth the line. We attempted a more complex example in Fig.~\ref{fig:cd_vs_sg}, where the robot's goal was to create a ripple in the sand. The effect of visual planning did not appear very strong here, even though quantitatively the visual guidance was supportive. We hypothesize that this was because it is very challenging to make smooth surfaces on sand, as even the lightest touch with the end-effector tends to make a strong visual indentation.

\vspace{-7pt}
\section{Discussions}

\subsection{Limitations}

A large technical limitation of our approach stems from our RGB+D sensor which is very expensive and fixed in a static location leading to a single perspective 3D view.
It would be possible to plan and model the dynamics of multiple perspectives to get a truly 3D sculpting method if the sensor could move to additional positions. Another limitation of our approach is the simple action parameters that are executed only from the top down. For a more expressive approach with more 3D capabilities, more complex actions could be designed.

\subsection{The Sensitivity of Visual Guidance to Noise}

\begin{figure}
    \centering
    \includegraphics[width=\linewidth]{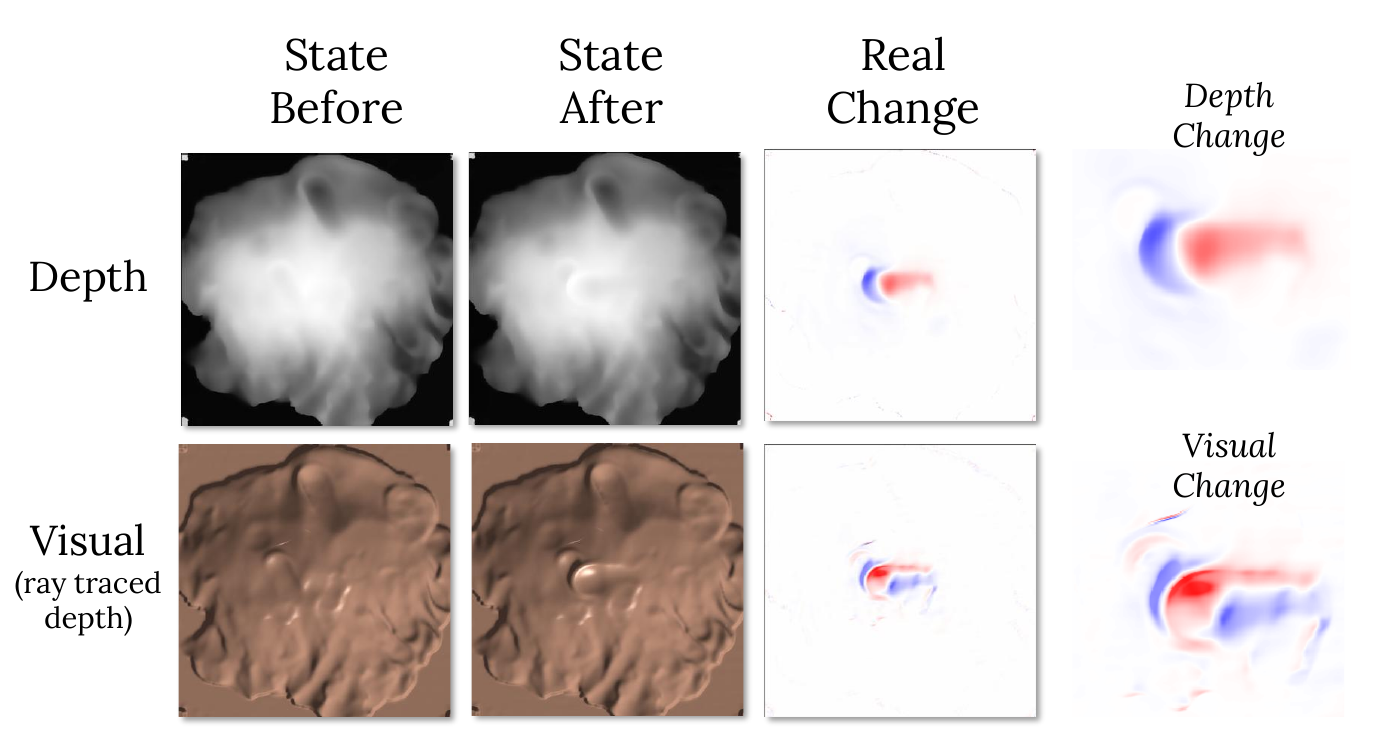} \vspace{-23pt}
    \caption{\textbf{Sensitivity of Visual Representations.} Depth maps are shown before and after an action is taken along with ray traced conversions of each. The changes in depth appear less complex than the change in ray traced images (averaged over RGB channels).}
    \label{fig:rgb_vs_depth_change} 
\end{figure}

In multiple experiments including dynamics modeling (Table~\ref{tab:dm_viz_3d}), long-horizon sculpting (Fig.~\ref{fig:losses-long-horizon}), and smoothing surfaces (Fig.~\ref{fig:cd_vs_sg}), we observed that our robot is successfully able to use visual guidance, in addition to 3D guidance. However, this visual guidance sometimes had little or no effect, as was seen with the sand example in Fig.~\ref{fig:cd_vs_sg}. We hypothesize that this is a result of the visually-aligned representation being highly sensitive to noise and less unpredictable than 3D representations. In Fig.~\ref{fig:rgb_vs_depth_change}, an action applied to a relatively flat surface had a simple change in depth, with a large indentation surrounded by some displaced material. However, in the ray-traced visual representation, this change is more complex. 

\begin{figure}[t!]
    \centering
    \includegraphics[width=\linewidth]{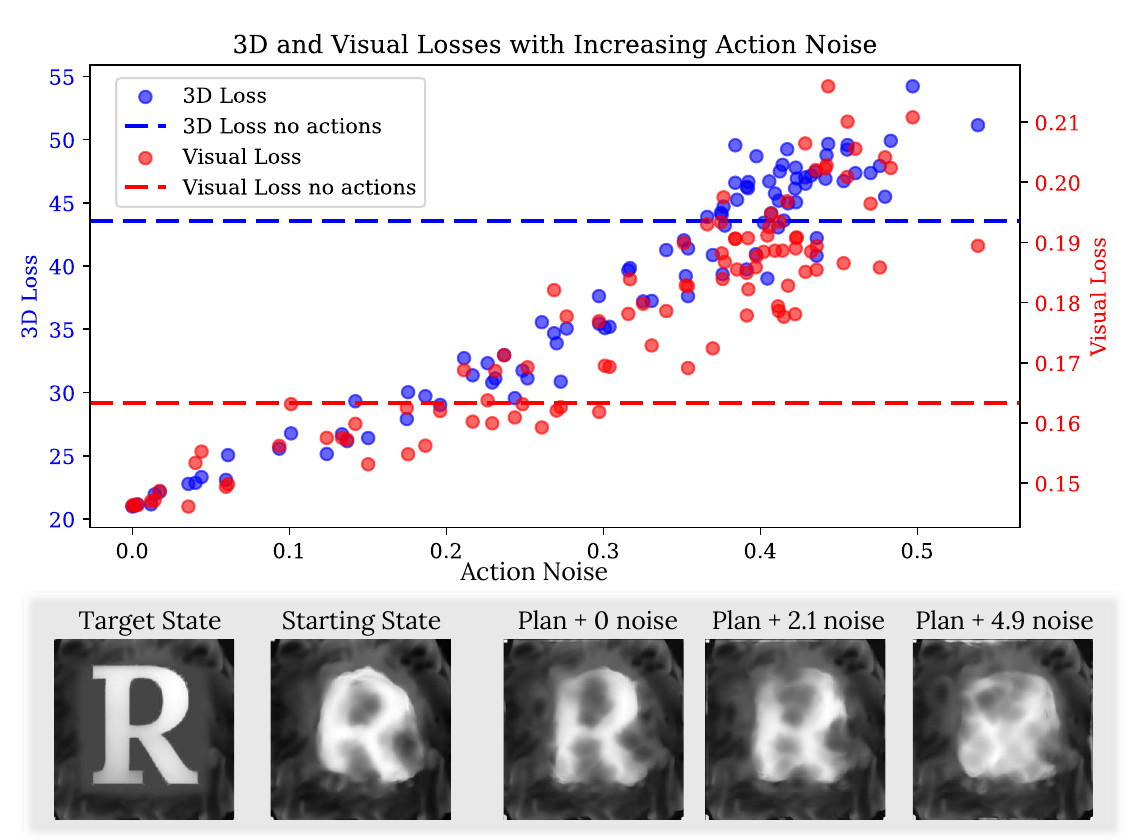} 
    \caption{\textbf{Noise versus Visual and 3D Accuracy.} Above, we plot the visual and 3D losses as more Gaussian noise is added to planned action parameters, simulating real-world noise. Samples of depth maps of plans with increasing noise added are shown below the plot.}
    \label{fig:noise_vs_loss} \vspace{-17pt}
\end{figure}

We hypothesize that the Sim2Real gap has a more drastic effect on visually-aligned representations compared to 3D representations. We simulated a Sim2Real gap by adding Gaussian noise to the action parameters of a planned sculpture (simulated output of the dynamics model).
In Fig.~\ref{fig:noise_vs_loss}, the losses were plotted as the amount of noise added to the actions increased. We also plotted the loss if zero actions were taken as horizontal, dashed lines. As expected, both the 3D and visual losses increased with additional added noise. However, the visual loss reached a point where the robot is better off not taking action with a smaller amount of noise than with 3D loss. 

In Fig.~\ref{fig:noise_vs_loss}, we measured the amount of noise as the MSE between the predicted and the noised actions which is the same calculation as $\mathcal{L}_{3D}$ in Table~\ref{tab:dm_viz_3d}.
The point where the robot is better off not performing any actions with respect to visual loss is roughly 0.2 in Fig.~\ref{fig:noise_vs_loss}. As seen in Table~\ref{tab:dm_viz_3d}, our Sim2Real gap is between 0.05 and 0.19, depending on the material, indicating that our method is barely able to perform visual planning without the Sim2Real gap being too high. This experiment supports our hypothesis that visual planning is very challenging due to the Sim2Real gap, but it is just one experiment conducted in simulation and more evidence is needed to make broader conclusions.

\vspace{-5pt}
\section{Conclusions}

We introduced a method for robotic deformable manipulation, tested on a variety of materials with multiple, diverse end-effectors. 
We showed that our pushing actions created smoother, more visually-aligned sculptures than the conventionally used pinching actions during long-horizon sculpting.
Through multiple experiments, we showed that using visually-aligned representations aided in both dynamics modeling and long-horizon planning. However, we also performed experiments providing evidence that visually-aligned representations are highly sensitive to a Sim2Real gap, making it a challenging modality to plan in for real-world robotics. While more experiments are necessary, we give evidence to the promise of planning robotic sculptures in  visual representations.

\footnotesize

\bibliographystyle{IEEEbib}
\bibliography{ref}

\end{document}